\documentclass[english]{article}
\usepackage[T1]{fontenc}
\usepackage[latin9]{inputenc}
\usepackage{geometry}
\usepackage[unicode=true,
 bookmarks=false,
 breaklinks=false,
 pdfborder={0 0 1},
 colorlinks=false]{hyperref}
\hypersetup{colorlinks,
linkcolor=red,
anchorcolor=blue,
citecolor=blue,
urlcolor=blue}
\usepackage{bm}
\usepackage{amsmath}
\usepackage{mathtools}
\usepackage{amssymb}
\usepackage{amsthm}
\usepackage{comment}
\usepackage{natbib}
\usepackage{booktabs}
\usepackage{graphicx}
\usepackage{algorithm}
\usepackage[noend]{algorithmic}
\usepackage{float}
\usepackage{multirow}
\usepackage{dsfont}
\usepackage{tcolorbox}
\usepackage{tabulary}
\usepackage{colortbl}
\usepackage{color}
\definecolor{gen}{RGB}{0,0,200}
\definecolor{cc}{RGB}{231,117,0}

\allowdisplaybreaks

\newcommand{\mymid}{\,|\,}

\title{Connections between reinforcement learning with feedback,  test-time scaling,  and diffusion guidance: An anthology}

\author{%
Yuchen Jiao\thanks{Department of Statistics and Data Science, Chinese University of Hong Kong.}
\and Yuxin Chen\thanks{Department of Statistics and Data Science, the Wharton School, University of Pennsylvania.} 
\and Gen Li\footnotemark[1] 
\thanks{Corresponding author}
}

\date{\today}

\begin{document}

\theoremstyle{plain} 
\newtheorem{lemma}{\bf Lemma} 
\newtheorem{proposition}{\bf Proposition}
\newtheorem{theorem}{\bf Theorem}
\newtheorem{corollary}{\bf Corollary} 
\newtheorem{claim}{\bf Claim}

\theoremstyle{remark}
\newtheorem{assumption}{\bf Assumption} 
\newtheorem{definition}{\bf Definition} 
\newtheorem{condition}{\bf Condition}
\newtheorem{property}{\bf Property} 
\newtheorem{example}{\bf Example}
\newtheorem{fact}{\bf Fact}
\newtheorem{remark}{\bf Remark}

\maketitle

\begin{abstract}
In this note, we reflect on several fundamental connections among widely used post-training techniques. 
We clarify some intimate connections and equivalences between reinforcement learning with human feedback, reinforcement learning with internal feedback, and test-time scaling (particularly soft best-of-$N$ sampling), while also illuminating  intrinsic links between diffusion guidance and test-time scaling. 
Additionally, we introduce a resampling approach for alignment and reward-directed diffusion models,  sidestepping the need for explicit reinforcement learning techniques. 
\end{abstract}

\noindent \textbf{Keywords:} RLHF, RLIF, test-time scaling, best-of-$N$ sampling, diffusion guidance, reward-directed diffusion models, importance sampling


\section{Introduction}

The rise of large language models (LLMs) has catalyzed a diverse array of post-training techniques that are reshaping the frontiers of artificial intelligence. 
Approaches such as reinforcement learning with feedback --- e.g.,  reinforcement learning with human feedback (RLHF) \citep{kaufmann2023survey} and reinforcement learning with internal feedback (RLIF) \citep{zhao2025learning} --- and 
test-time scaling methods (e.g., best-of-$N$ sampling) \citep{snell2024scaling} have become pivotal in enhancing model performance. 
Historically, these approaches have evolved along nearly independent trajectories, with their mathematical underpinnings developed largely in isolation.  
This technical note reflects on some basic, yet intricate, connections among these paradigms, offering some unified  perspectives that might inform and improve the design of each. 

\paragraph{An overview of our results.} 
In this short note, we discuss the intrinsic connections among the post-training techniques discussed above. Our findings are summarized below, tailored to both autoregressive and diffusion models. 
\begin{itemize}
    \item[1)] We demonstrate that, RLHF and RLIF can be formulated equivalently, assuming suitable parameter choices and  faithfulness of reference policies.

    \item[2)] We show that soft best-of-$N$ sampling for test-time scaling is asymptotically (as $N\rightarrow \infty$) equivalent to RLHF, and is also equivalent to RLIF under further assumptions on the reference policy.   

    \item[3)] We propose an alternative to RLHF and RLIF for alignment, dubbed as ``RL-free alignment.'' The proposed technique attempts maximum likelihood estimation with the aid of importance sampling, sidestepping the need for explicit RL techniques.


    \item[4)] Further, we point out that soft best-of-$N$ sampling can be applied to diffusion models to achieve the goal of classifier-free diffusion guidance, and the above resampling approach can be extended to tackle reward-directed diffusion models without resorting to RL-based techniques.  

    
\end{itemize}


\paragraph{Notation.} 
For any distribution $p$
supported on $\mathcal{V}$, its  entropy is defined as
\begin{align}
\label{eq:defn-entropy}
\mathcal{H}(p)\coloneqq -\sum_{v \in \mathcal{V}} p(v)\log(p(v)). 
\end{align}
For any two distributions $p$ and $q$ supported on $\mathcal{V}$, we denote by $\mathsf{KL}(p \parallel q) = \sum_{x\in \mathcal{V}} p(v)\log \frac{p(v)}{q(v)}$ the Kullback-Leibler (KL) divergence from $q$ to $p$. More generally, when $p$ and $q$ are non-negative vectors (but not necessarily probability vectors), we define the generalized KL divergence as
\begin{align}
\mathsf{KL}(p \parallel q) \coloneqq \sum_{v\in \mathcal{V}} \left\{ p(v)\log \frac{p(v)}{q(v)} - p(v) + q(v) \right\}, 
\label{eq:def-KL-general}
\end{align}
which is the Bregman divergence generated by the negative entropy  \citep{beck2017first}. 
For  two vectors $a,b\in \mathbb{R}^n$ and scalar $\alpha$, denote $a^{\alpha}= [a_i^{\beta}]_{1\leq i\leq n}$ and $ab=[a_ib_i]_{1\leq i\leq n}$; namely, these operations are applied entriwise.

\section{Autoregressive models}
\label{sec:AR-model}

We begin by exploring connections under the autoregressive models, which are foundational statistical models underlying next-token prediction in LLMs.

Consider the task of generating an answer $o$ --- taking the form of a random output sequence --- given query $q$. 
At a high level, the AR model assumes that each token's generation depends on all previously generated tokens.  Given that our results do not rely on any specific formulae of the AR model, we omit detailed formulation here but instead refer the interested reader to standard references like \citep{jm3}. 
Throughout this section, we assume access to a reference distribution or a reference policy, denoted by $\pi_{\mathsf{ref}}$ throughout, typically obtained during the pre-training or fine-tuning stage.

\subsection{Preliminaries}

\paragraph{Reinforcement learning with human feedback.}  RLHF, which harnesses human preferences to align LLMs with human values, 
has gained widespread adoption in LLMs \citep{ouyang2022training,kaufmann2023survey}. More often than not, the human preferences are gathered through pairwise comparisons, which are then used to train a reward function that reflects human values and steers the LLM's responses. 

Setting the stage, consider two candidate answers, $o$ and $\widetilde{o}$, generated by an LLM in response to query $q$. A human evaluator is tasked with determining whether $o$ is preferred over $\widetilde{o}$ or vice versa. The  comparison outcome is frequently modeled using the classical the Bradley-Terry (BT) model \citep{bradley1952rank,chen2021spectral}, which assumes the probability of $o$ being preferred over $\widetilde{o}$ (denoted by $o\succ \widetilde{o}$) to be
\begin{align}
\mathbb{P}(o \succ \widetilde{o} \mymid q) = \frac{\exp(r(q, o))}{\exp(r(q, o)) + \exp(r(q, \widetilde{o}))},
\label{eq:BT-model}
\end{align}
where $r$ is some unknown reward function mapping any query-answer pair to a scalar value. 
In light of this BT model, we can readily introduce a human preference distribution, denoted by, $\pi_{\mathsf{hp}}$, such that
\begin{align}
\pi_{\mathsf{hp}}(o\mymid q)\coloneqq\frac{\exp(r(q, o))}{\sum_o \exp(r(q, o))}.
\label{eq:defn-pi-HP}
\end{align}
This distribution captures human preferences, assuming that the BT model \eqref{eq:BT-model} aligns with human values.




In RLHF, a standard approach typically involves two key steps. 
Firstly, the collected human feedback is utilized to train a reward model --- denoted by $r_{\phi}$ with parameterization $\phi$ --- either explicitly or implicitly. Secondly, one attempts to derive a policy aimed at maximizing the learned reward, while in the meantime ensuring that the derived policy does not deviate much from the reference policy $\pi_{\mathsf{ref}}$. For a given query $q$, this can be formulated through the following regularized optimization objective: 
\begin{align}\label{eq:obj-RLHF-0}
\mathop{\text{maximize}}\limits_{\theta} ~~ \mathop{\mathbb{E}}\limits_{o \sim \pi_{\theta}(\cdot\mymid q)}[r_{\phi}(q, o)] - \beta_{\mathsf{hf}}\mathsf{KL}\big(\pi_{\theta}(\cdot \mymid q) \parallel \pi_{\mathsf{ref}}(\cdot \mymid q) \big).
\end{align}
Here, policy estimate $\pi_{\theta}$ is parameterized by $\theta$, and $\beta_{\mathsf{hf}}>0$ represents some regularization parameter.




\paragraph{Reinforcement learning with internal feedback.} 
The RLIF framework bypasses the step of collecting additional human feedback to train an external reward function. Instead, it utilizes a self-certainty metric as an internal reward function. 
The original version of this metric is defined as \citep{kang2025scalable}
\begin{align}
\label{eq:defn-self-certainty}
u^{\mathsf{orig}}(q, o) \coloneqq \frac{1}{L}\sum_{i = 1}^{L}\mathsf{KL}\big(\mathsf{Unif}(\mathcal{V}) \parallel \pi_{\theta}(\cdot\mymid q, o_{<i}) \big),
\end{align}
where $L$ is the length of the response $o$, 
$\mathsf{Unif}(\mathcal{V})$ denotes uniform distribution  over the vocabulary $\mathcal{V}$,   and $\pi_{\theta}(\cdot\mymid q, o_{<i})$ represents the model's predicted distribution given query $q$ and all output prior to token $i$.
The metric was originally motivated by the empirical observation that: LLMs tend to exhibit lower confidence when faced with unfamiliar tasks or insufficient knowledge \citep{kang2024unfamiliar}; 
 conversely, higher self-certainty scores --- which indicate more confident predictions --- are often associated  with more reliable responses.

In this note, we consider a closely related variation of the original self-certainty metric \eqref{eq:defn-self-certainty} as follows:
\begin{align*}
u(q, o) \coloneqq \frac{1}{L}\sum_{i = 1}^{L}\mathsf{KL}\big(\pi_{\theta}(\cdot\mymid q, o_{<i}) \parallel \mathsf{Unif}(\mathcal{V}) \big)
&= \frac{1}{L}\sum_{i = 1}^{L}\sum_{v\in \mathcal{V}}\pi_{\theta}(v\mymid q, o_{<i})\log\big(|\mathcal{V}|\pi_{\theta}(v\mymid q, o_{<i})\big),
\end{align*}
and take the expected self-certainty as the internal reward:
\begin{align*}
\mathop{\mathbb{E}}\limits_{o \sim \pi_{\theta}(\cdot\mymid q)}[u(q, o)]
= \log(|\mathcal{V}|) + \frac{1}{L}\sum_{o }\pi_{\theta}(o\mymid q)\log\big(\pi_{\theta}(o\mymid q)\big) = \log(|\mathcal{V}|) - \frac{1}{L}\mathcal{H}(\pi_{\theta}(\cdot\mymid q)),
\end{align*}
where $\mathcal{H}(\pi_{\theta}(\cdot\mymid q))$ denotes the entropy of $\pi_{\theta}(\cdot\mymid q)$ (see \eqref{eq:defn-entropy}).
With this choice in place, the RLIF optimization objective can be expressed as the following KL-regularized problem: 
\begin{align}
\mathop{\text{maximize}}\limits_{\theta} \mathop{\mathbb{E}}\limits_{o \sim \pi_{\theta}(\cdot\mymid q)}[u_{\theta}(q, o)] - \beta_{\mathsf{if}}\mathsf{KL}(\pi_{\theta}(\cdot\mymid q) \parallel \pi_{\mathsf{ref}}(\cdot\mymid q))
\label{eq:formulation-RLIF},
\end{align}
where $\beta_{\mathsf{if}}>0$ is a regularization parameter. Here, we adopt the notation $u_{\theta}$ since self-certainty might depend on the parameterization $\theta$ of policy estimate $\pi_{\theta}$.

\paragraph{Test-time scaling} refers to a post-training strategy to enhance performance during inference without model retraining \citep{brown2024large,snell2024scaling,muennighoff2025s1}.
Among various approaches, the best-of-$N$ sampling method and its variations have recently received growing attention. In essence, for a given query $q$, 
the best-of-$N$ algorithms 
prompts the LLM to generate 
$N$ independent responses 
\begin{align}
o^{(1)}, o^{(2)}, \ldots, o^{(N)} \overset{\mathrm{i.i.d.}}{\sim} \pi_{\mathsf{ref}}(\cdot\mymid q),
\label{eq:N-ind-responses}
\end{align}
from which the most favorable response, denoted by $o^{\mathsf{best}}$, 
is selected according to some specific evaluation criterion (e.g., evaluated based on some derived reward function).

\subsection{Connections between RLHF, RLIF and test-time scaling}
\label{sec:AR-RL-tts}

This subsection  illuminates some intrinsic connections between RLHF, RLIF and test-time scaling.  In the sequel, we assume that the learned reward is accurate in the sense that $r_\phi = r$. 
To ease presentation, the proofs of the results in this subsection are postponed to Section~\ref{sec:proof-AR-model}.  

\paragraph{Connection between RLHF and RLIF.} 
We first point out some equivalent formulations of RLHF and RLIF under certain conditions. 
\begin{itemize}
\item {\em Equivalent objective of RLHF.}
When $r_{\phi}=r$, the RLHF problem \eqref{eq:obj-RLHF-0} can be alternatively expressed as
\begin{align}\label{eq:obj-RLHF}
&\arg\max_{\theta} \left\{ \mathop{\mathbb{E}}\limits_{o \sim \pi_{\theta}(\cdot|q)}[r(q, o)] - \beta_{\mathsf{hf}}\mathsf{KL}\big(\pi_{\theta}(\cdot \mymid q) \parallel \pi_{\mathsf{ref}}(\cdot\mymid q)\big) \right\} \notag\\
&\qquad = \arg\min_{\theta} \mathsf{KL}\big(\pi_{\theta}(\cdot\mymid q) \parallel \pi_{\mathsf{hp}}(\cdot\mymid q)^{1/\beta_{\mathsf{hf}}}\pi_{\mathsf{ref}}(\cdot\mymid q)\big),
\end{align}
where we recall the definitions of the generalized KL divergence and $\pi_{\mathsf{hp}}$ in \eqref{eq:def-KL-general} and \eqref{eq:defn-pi-HP}, respectively.  
In other words, the RLHF objective \eqref{eq:obj-RLHF-0} essentially leverages the reward function for exponential tilting, adjusting the reference policy by $\pi_{\mathsf{hp}}(\cdot\mymid q)^{1/\beta_{\mathsf{hf}}}$, or equivalently,  $\exp(r(q,\cdot)/\beta_{\mathsf{hf}})$.   Note that the exponential tilting feature of RLHF was  recognized in prior work (e.g., \citet{ziegler2019fine}).

\item {\em Equivalent objective of RLIF.}
 The RLIF objective \eqref{eq:formulation-RLIF} also admits an equivalent objective as follows:
\begin{align}\label{eq:obj-RLIF}
& \arg\max_{\theta} \left\{ \mathop{\mathbb{E}}\limits_{o \sim \pi_{\theta}(\cdot\mymid q)}[u_{\theta}(q, o)] - \beta_{\mathsf{if}}\mathsf{KL}\big(\pi_{\theta}(\cdot\mymid q) \parallel \pi_{\mathsf{ref}}(\cdot|q) \big) \right\} = \arg\min_{\theta} \mathsf{KL}\Big(\pi_{\theta}(\cdot\mymid q) \parallel \pi_{\mathsf{ref}}^{\frac{L\beta_{\mathsf{if}}}{L\beta_{\mathsf{if}}-1}}(\cdot\mymid q)\Big),
\end{align}
provided that $L\beta_{\mathsf{if}} > 1$. 

\end{itemize}

%
%

Now, consider the special case where the learned reference policy $\pi_{\mathsf{ref}}$ is aligned with human values in the sense that $\pi_{\mathsf{ref}}=\pi_{\mathsf{hp}}$.  Comparing \eqref{eq:obj-RLIF} with \eqref{eq:obj-RLHF} reveals that:  RLIF is equivalent to RLHF when the regularization parameters satisfy
\begin{align}\beta_{\mathsf{hf}} = L\beta_{\mathsf{if}}-1.
\label{eq:beta-equiv-RLHF-RLIF}
\end{align}
In this case, a unified optimization objective for both RLHF and RLIF is given by
\begin{align}\label{eq:obj-RLHF-IF}
\arg\min_{\theta} \mathcal{H}(\pi_{\theta}(\cdot\mymid q)) + (\beta_{\mathsf{hf}}+1)\mathsf{KL}\big(\pi_{\theta}(\cdot\mymid q) \parallel \pi_{\mathsf{hp}}(\cdot\mymid q)\big) 
&= \arg\min_{\theta} \mathsf{KL}\Big(\pi_{\theta}(\cdot\mymid q) \parallel \pi_{\mathsf{hp}}^{1+\beta_{\mathsf{hf}}^{-1}}(\cdot\mymid q)\Big).
\end{align}



\paragraph{Connection between test-time scaling and RL from feedback.}

Next, we single out an intimate connection between test-time scaling and RLHF. 
Specifically, 
consider a ``soft'' variation of the best-of-$N$ sampling algorithm (see \citet{verdun2025soft}), so that 
the final output $o^{\mathsf{best}}$ is selected stochastically
as 
\begin{align}
\label{eq:stochastic-best-of-N}
\mathbb{P}(o^{\mathsf{best}} = o^{(n)}\mymid q, o^{(1)}, o^{(2)}, \ldots, o^{(N)}) \propto \pi_{\mathsf{hp}}^{\beta_{\mathsf{tts}}^{-1}}(o^{(n)}\mymid q),
\end{align}  
with $o^{(1)},\dots,o^{(N)}$ the $N$ candidate responses.  Here, $\beta_{\mathsf{tts}}\geq 0$ is a hyperparameter; in particular, as $\beta_{\mathsf{tts}}\rightarrow 0$, the soft sampling scheme \eqref{eq:stochastic-best-of-N} reduces to the vanilla best-of-$N$ sampling evaluated based on the true reward $r$, namely, 
$$o^{\mathsf{best}}=\arg\max_{o\in \{o^{(1)},\dots,o^{(N)}\}}\pi_{\mathsf{hp}}(o\mymid q)
=\arg\max_{o\in \{o^{(1)},\dots,o^{(N)}\}}r(q,o)
\qquad (\text{when }\beta_{\mathsf{tts}}\rightarrow 0).
$$
%
As it turns out, this test-time scaling approach bears similarity with RLHF and RLIF, as explained below. 
\begin{itemize}
\item 
Assume that the reward function obeys 
 \begin{align}
 \label{eq:assumption-rmax}
 0\le r(q,o)\le r_{\mathsf{max}}<\infty, \qquad \text{for all }q\text{ and }o. 
 \end{align}
Then for any $q$ and $o$, the response $o^{\mathsf{best}}$ returned by \eqref{eq:stochastic-best-of-N} satisfies
\begin{align}
\label{eq:equiv-tts-RLHFz}
\mathbb{P}(o^{\mathsf{best}} = o\mymid q) \overset{N\rightarrow \infty}{\longrightarrow} 
\frac{\pi_{\mathsf{ref}}(o\mymid q)\pi_{\mathsf{hp}}^{\beta_{\mathsf{tts}}^{-1}}(o\mymid q)}{\sum_{o} \pi_{\mathsf{ref}}(o\mymid q)\pi_{\mathsf{hp}}^{\beta_{\mathsf{tts}}^{-1}}(o\mymid q)}.
\end{align}
Comparing \eqref{eq:equiv-tts-RLHFz} with \eqref{eq:obj-RLHF} unveils a curious fact: soft best-of-$N$ sampling is equivalent to the optimal unconstrained distribution obtained in RLHF, provided the regularization parameters obey 
\begin{align}
\beta_{\mathsf{tts}} = \beta_{\mathsf{hf}}.
\end{align}
 %

\item 
Furthermore, the above result combined with \eqref{eq:beta-equiv-RLHF-RLIF} reveals that: 
in the idealistic case where  $\pi_{\mathsf{ref}} = \pi_{\mathsf{hp}}$, 
the output distribution of the soft best-of-$N$ sampling method \eqref{eq:stochastic-best-of-N} simplifies to
\begin{align}
\label{eq:equiv-tts-RLIF}
\mathbb{P}(o^{\mathsf{best}} = o\mymid q) \overset{N\rightarrow \infty}{\longrightarrow} 
\frac{\pi_{\mathsf{ref}}^{1+\beta_{\mathsf{tts}}^{-1}}(o\mymid q)}{\sum_o \pi_{\mathsf{ref}}^{1+\beta_{\mathsf{tts}}^{-1}}(o\mymid q)},
\end{align}
which coincides with the optimal unconstrained distribution of RLIF (cf.~\eqref{eq:obj-RLIF})  
as long as
\begin{align}
\beta_{\mathsf{tts}} 
= L\beta_{\mathsf{if}}-1.
\end{align}
%

\end{itemize}



\subsection{An alternative resampling approach:  RL-free alignment}
\label{sec:resampling-AR}

%


Motivated by the  preliminary facts about RLHF and RLIF, 
we put forward an alternative approach to alignment, which we refer to as ``RL-free alignment.''  

To begin with, 
 recall from \eqref{eq:obj-RLHF} and \eqref{eq:defn-pi-HP} that RLHF can be interpreted as a procedure that tilts the reference policy  $\pi_{\mathsf{ref}}(\cdot \mymid q)$ to $\exp(r(q, \cdot)/\beta_{\mathsf{hf}})\pi_{\mathsf{ref}}(\cdot\mymid q)$ with suitable normalization, where $r$ is the reward function derived based on human preferences.  
This motivates direct alignment of a model towards the adjusted distribution without explicitly invoking RL techniques. 
Specifically, we propose to approximate the exponentially-tilted distribution 
$\frac{\exp(r(q, o)/\beta_{\mathsf{hf}})\pi_{\mathsf{ref}}(o\mymid q)}{\sum_{o'} \exp(r(q, o')/\beta_{\mathsf{hf}})\pi_{\mathsf{ref}}(o'\mymid q)}$
by solving the following problem:
%
\begin{align}\label{eq:fintunenoRL-AR}
\mathop{\text{maximize}}\limits_{\theta} \quad Z_{\mathsf{hf}}^{-1}(q)\mathop{\mathbb{E}}\limits_{o \sim \pi_{\mathsf{ref}}(\cdot\mymid q)}\big[\exp\big(r(q, o)/\beta_{\mathsf{hf}}\big)\log \pi_{\theta}(o\mymid q) \big],
\end{align}
where the normalized constant $Z_{\mathsf{hf}}(q)$ is given by 
\begin{align*}
Z_{\mathsf{hf}}(q) = \mathop{\mathbb{E}}\limits_{o \sim \pi_{\mathsf{ref}}(\cdot\mymid q)}\big[\exp\big(r(q, o)/\beta_{\mathsf{hf}}\big)\big].
\end{align*}
To explain the rationale, a common strategy to approximate a target distribution $p_{\mathsf{target}}$  involves solving
\[
\arg\min_{\theta} ~~\mathsf{KL}\big( p_{\mathsf{target}}\parallel \pi_{\theta}\big)
\qquad \Longleftrightarrow \qquad 
\arg\max_{\theta} ~~ \sum_x p_{\mathsf{target}}(x) \log \pi_{\theta}(x). 
\]
%
Further, it is readily seen that 
the objective \eqref{eq:fintunenoRL-AR} is equivalent to maximizing the expected log-likelihood under an adjusted data distribution; that is, 
\begin{align}
&\arg\max_{\theta} Z_{\mathsf{hf}}^{-1}(q) \mathop{\mathbb{E}}\limits_{o \sim \pi_{\mathsf{ref}}(\cdot\mymid q)}\big[\exp(r(q, o)/\beta_{\mathsf{hf}})\log \pi_{\theta}(o\mymid q)\big] 
\notag\\
&\qquad =
\arg\max_{\theta} \mathop{\mathbb{E}}\limits_{o \sim Z_{\mathsf{hf}}^{-1}(q)\exp(r(q, \cdot)/\beta_{\mathsf{hf}})\pi_{\mathsf{ref}}(\cdot\mymid q)}\big[\log \pi_{\theta}(o\mymid q)\big]. 
\label{eq:resampling-RLHF}
\end{align}
%
In light of \eqref{eq:resampling-RLHF}, we propose to compute the maximum likelihood estimation (MLE) with respect to an adjusted distribution proportional to 
$\exp(r(q, \cdot)/\beta_{\mathsf{hf}})\pi_{\mathsf{ref}}(\cdot\mymid q)$. 
In practice, the expectation operator can be replaced by a finite-sample approximation, where the data samples are independently drawn from $o\sim \pi_{\mathsf{ref}}(\cdot\mymid q)$ and importance sampling can be invoked to adjust the effective data distribution to be proportional to  $\exp(r(q, \cdot)/\beta_{\mathsf{hf}})\pi_{\mathsf{ref}}(\cdot\mymid q)$. 
Crucially, the proposed approach follows the MLE paradigm in conjunction with importance sampling, without explicitly resorting to RL-based methods. 

Akin to RLHF, 
we also come up with an RL-free approach to achieving similar goals as RLIF. 
More concretely, 
we propose to  approximate
the distribution proportional to $\pi_{\mathsf{ref}}^{\frac{L\beta_{\mathsf{if}}}{L\beta_{\mathsf{if}}-1}}(\cdot\mymid q)$ by solving:
\begin{align}\label{eq:fintunenoRLIF-AR}
\mathop{\text{maximize}}\limits_{\theta} \quad Z_{\mathsf{if}}^{-1}(q)\mathop{\mathbb{E}}\limits_{o \sim \pi_{\mathsf{ref}}(\cdot\mymid q)}\big[\pi_{\mathsf{ref}}^{\frac{1}{L\beta_{\mathsf{if}}-1}}(o\mymid q)\log \pi_{\theta}(o\mymid q) \big],
\end{align}
where the normalized constant 
is
$Z_{\mathsf{if}}(q) = \mathop{\mathbb{E}}\limits_{o \sim \pi_{\mathsf{ref}}(\cdot\mymid q)}\big[\pi_{\mathsf{ref}}^{\frac{1}{L\beta_{\mathsf{if}}-1}}(o\mymid q)\big].$
Invoking a similar argument tells us that: this objective \eqref{eq:fintunenoRLIF-AR} is equivalent to computing the following MLE w.r.t.~an
adjusted distribution:
\begin{align}
&\arg\max_{\theta} Z_{\mathsf{if}}^{-1}(q) \mathop{\mathbb{E}}\limits_{o \sim \pi_{\mathsf{ref}}(\cdot\mymid q)}\big[\pi_{\mathsf{ref}}^{\frac{1}{L\beta_{\mathsf{if}}-1}}(o\mymid q)\log \pi_{\theta}(o\mymid q)\big] 
\notag\\
&\qquad =
\arg\max_{\theta} \mathop{\mathbb{E}}\limits_{o \sim Z_{\mathsf{if}}^{-1}(q)\pi_{\mathsf{ref}}^{\frac{L\beta_{\mathsf{if}}}{L\beta_{\mathsf{if}}-1}}(\cdot\mymid q)}\big[\log \pi_{\theta}(o\mymid q)\big]. 
\label{eq:resampling-RLIF}
\end{align}
As previously, we can draw data samples from $\pi_{\mathsf{ref}}$ and apply importance sampling paradigms to approximate the expection in \eqref{eq:resampling-RLIF}, thereby bypassing the need of RL techniques.

\subsection{Proofs}
\label{sec:proof-AR-model}

\paragraph{Proof of relation~\eqref{eq:obj-RLHF}.} Regarding the RLHF objective, we make the observation that
\begin{align}
&\arg\max_{\theta} \left\{ \mathop{\mathbb{E}}\limits_{o \sim \pi_{\theta}(\cdot|q)}[r(q, o)] - \beta_{\mathsf{hf}}\mathsf{KL}\big(\pi_{\theta}(\cdot \mymid q) \parallel \pi_{\mathsf{ref}}(\cdot\mymid q)\big) \right\} \notag\\
&\qquad=\arg\min_{\theta} \left\{ -\frac{1}{\beta_{\mathsf{hf}}}\mathop{\mathbb{E}}\limits_{o \sim \pi_{\theta}(\cdot|q)}[r(q, o)] +\mathsf{KL}\big(\pi_{\theta}(\cdot\mymid q) \parallel \pi_{\mathsf{ref}}(\cdot\mymid q)\big) \right\} \notag\\
&\qquad=\arg\min_{\theta}\sum_o \left(\pi_{\theta}(o\mymid q)\log\frac{\pi_{\theta}(o\mymid q)}{\pi_{\mathsf{ref}}(o\mymid q)} - \frac{1}{\beta_{\mathsf{hf}}}\pi_{\theta}(o\mymid q)r(q, o)\right)\notag\\
&\qquad=\arg\min_{\theta}\sum_o \pi_{\theta}(o\mymid q)\log\frac{\pi_{\theta}(o\mymid q)}{\exp(r(q,o)/\beta_{\mathsf{hf}})\pi_{\mathsf{ref}}(o\mymid q)}\notag\\
&\qquad = \arg\min_{\theta} \left\{ \sum_o \pi_{\theta}(o\mymid q)\log\frac{\pi_{\theta}(o\mymid q)}{\pi_{\mathsf{hp}}(o\mymid q)^{1/\beta_{\mathsf{hf}}}
\pi_{\mathsf{ref}}(o\mymid q)} 
-  \sum_o \pi_{\theta}(o\mymid q)
\log \Big(\sum_o \exp\big(r(q,o)\big)\Big)^{\frac{1}{\beta_{\mathsf{hf}}}} \right\}
\notag\\
&\qquad \overset{\mathrm{(i)}}{=}  \arg\min_{\theta}\sum_o \pi_{\theta}(o\mymid q)\log\frac{\pi_{\theta}(o\mymid q)}{\pi_{\mathsf{hp}}(o\mymid q)^{1/\beta_{\mathsf{hf}}}
\pi_{\mathsf{ref}}(o\mymid q)} 
\notag\\
&\qquad \overset{\mathrm{(ii)}}{=} \arg\min_{\theta}\sum_o \pi_{\theta}(o\mymid q)\log\frac{\pi_{\theta}(o\mymid q)}{\pi_{\mathsf{hp}}(o\mymid q)^{1/\beta_{\mathsf{hf}}}
\pi_{\mathsf{ref}}(o\mymid q)} 
- \sum_o \pi_{\theta}(o\mymid q) + \sum_o \log \big( \pi_{\mathsf{hp}}(o\mymid q)^{1/\beta_{\mathsf{hf}}}\pi_{\mathsf{ref}}(o\mymid q) \big)\notag\\
&\qquad\overset{\mathrm{(iii)}}{=} \arg\min_{\theta} \mathsf{KL}\big(\pi_{\theta}(\cdot\mymid q) \parallel \pi_{\mathsf{hp}}(\cdot\mymid q)^{1/\beta_{\mathsf{hf}}}\pi_{\mathsf{ref}}(\cdot\mymid q)\big),
\end{align}
where (i) and (ii) are valid since both 
$\log \big(\sum_o \exp(r(q,o))\big)$ and 
$\sum_o \log \big( \exp(r(q,o)/\beta)\pi_{\mathsf{ref}}(o\mymid q) \big)$ are independent of $\theta$ and $\sum_o \pi_{\theta}(o\mymid q)=1$, and (iii) follows from the definitions \eqref{eq:def-KL-general} and \eqref{eq:defn-pi-HP}.

\paragraph{Proof of relation~\eqref{eq:obj-RLIF}.}

Regarding the RLIF objective, one has
\begin{align}
& \arg\max_{\theta} \left\{ \mathop{\mathbb{E}}\limits_{o \sim \pi_{\theta}(\cdot\mymid q)}[u_{\theta}(q, o)] - \beta_{\mathsf{if}}\mathsf{KL}\big(\pi_{\theta}(\cdot\mymid q) \parallel \pi_{\mathsf{ref}}(\cdot|q) \big) \right\}\notag\\
&\qquad = \arg\min_{\theta} \left\{\mathcal{H}\big(\pi_{\theta}(\cdot\mymid q)\big) + L\beta_{\mathsf{if}}\mathsf{KL}\big(\pi_{\theta}(\cdot\mymid q) \parallel \pi_{\mathsf{ref}}(\cdot\mymid q) \big) 
\right\} 
\notag\\
&\qquad =
\arg\min_{\theta}
\left\{
-\sum_o \pi_{\theta}(o\mymid q) \log \big( \pi_{\theta}(o\mymid q) \big) 
+ L\beta_{\mathsf{if}} 
\sum_o \pi_{\theta}(o\mymid q) \log  \frac{\pi_{\theta}(o\mymid q) }{\pi_{\mathsf{ref}}(o\mymid q)} 
\right\}
\notag\\
&\qquad =
\arg\min_{\theta}
\left\{ (L\beta_{\mathsf{if}} -1)\sum_o \pi_{\theta}(o\mymid q) \log\frac{\pi_{\theta}(o\mymid q)}{\left(\pi_{\mathsf{ref}}(o\mymid q)\right)^{\frac{L\beta_{\mathsf{if}}}{L\beta_{\mathsf{if}}-1}}} 
\right\} 
\notag\\
&\qquad = \arg\min_{\theta} \mathsf{KL}\Big(\pi_{\theta}(\cdot\mymid q) \parallel \pi_{\mathsf{ref}}^{\frac{L\beta_{\mathsf{if}}}{L\beta_{\mathsf{if}}-1}}(\cdot\mymid q)\Big),
\end{align}

 \paragraph{Proof of relation~\eqref{eq:equiv-tts-RLHFz}.} 
Before proceeding, note that  
in view of the assumption~\eqref{eq:assumption-rmax} and the definition \eqref{eq:defn-pi-HP} of $\pi_{\mathsf{hp}}$,  there exist $\pi_{\mathsf{min}}$ and $\pi_{\mathsf{max}}$  such that
 \begin{align}
 \label{eq:best-of-N-assumption-pi}
 \pi_{\mathsf{min}} \le 
 \pi_{\mathsf{hp}}(o\mymid q)\le \pi_{\mathsf{max}} ~\text{ for any }q\text{ and }o,\qquad \mathrm{and} ~~\frac{\pi_{\mathsf{max}}}{\pi_{\mathsf{min}}}\le \exp(r_{\mathsf{max}}).
 \end{align}

Recall that $o^{(1)},\dots,o^{(N)}$ are 
independently drawn from $\pi_{\mathsf{ref}}(\cdot\mymid q)$ (see \eqref{eq:N-ind-responses}).  
Under the soft best-of-$N$ sampling in \eqref{eq:stochastic-best-of-N}, the distribution of $o^{\mathsf{best}}$ can be calculated as
\begin{align}\label{eq:testtimescaling}
\mathbb{P}(o^{\mathsf{best}} = o\mymid q) 
&= \sum_{n=1}^N \mathbb{P}\big(o^{\mathsf{best}} = o^{(n)},o^{(n)} = o\mymid q\big)\notag\\
&=\sum_{n = 1}^N \mathbb{E}_{\{o^{(i)}:i\neq n\}}\Big[\mathbb{P}\big(o^{\mathsf{best}} = o^{(n)}\mymid q, o^{(n)} = o, \{o^{(i)}:i\neq n\}\big)\mathbb{P}(o^{(n)} = o\mymid q) \,\Big|\,q\Big] \notag\\
&=\sum_{n=1}^N\mathbb{E}_{\{o^{(i)}:i\neq n\}}\left[\frac{\pi_{\mathsf{ref}}(o\mymid q)\pi_{\mathsf{hp}}^{\beta_{\mathsf{tts}}^{-1}}(o\mymid q)}{\pi_{\mathsf{hp}}^{\beta_{\mathsf{tts}}^{-1}}(o\mymid q) + \sum_{i: i\ne n} \pi_{\mathsf{hp}}^{\beta_{\mathsf{tts}}^{-1}}(o^{(i)}\mymid q)} \,\bigg|\,q \right]\notag\\
&=N\mathbb{E}_{\{o^{(i)}:i\ne 1\}}\left[\frac{\pi_{\mathsf{ref}}(o\mymid q)\pi_{\mathsf{hp}}^{\beta_{\mathsf{tts}}^{-1}}(o\mymid q)}{\pi_{\mathsf{hp}}^{\beta_{\mathsf{tts}}^{-1}}(o\mymid q) + \sum_{i: i\ne 1} \pi_{\mathsf{hp}}^{\beta_{\mathsf{tts}}^{-1}}(o^{(i)}\mymid q)}\,\bigg|\,q\right],
\end{align}
where the last identity invokes the fact that $\{o^{(i)}\}$ are i.i.d.~drawn.

Equipped with Assumption~\eqref{eq:best-of-N-assumption-pi}, 
applying the Bernstein inequality reveals that
\begin{align}\label{eq:bernstein}
 & \bigg|\frac{1}{N-1}\sum_{i\ne 1}\pi_{\mathsf{hp}}^{\beta_{\mathsf{tts}}^{-1}}(o^{(i)}\mymid q)-\sum_{o}\pi_{\mathsf{ref}}(o\mymid q)\pi_{\mathsf{hp}}^{\beta_{\mathsf{tts}}^{-1}}(o\mymid q)\bigg|^{2}\notag\\
 & \qquad = O\left(\frac{\log N}{N}\mathop{\mathbb{E}}\limits _{o\sim\pi_{\mathsf{ref}}(\cdot\mymid q)}\left[\big(\pi_{\mathsf{hp}}^{\beta_{\mathsf{tts}}^{-1}}(o\mymid q)\big)^{2}\right]+\frac{(\pi_{\mathsf{max}}^{\beta_{\mathsf{tts}}^{-1}})^{2}\log^{2}N}{N^{2}}\right)\notag\\
 & \qquad\leq O\left(\frac{\log N}{N}\sum_{o}\pi_{\mathsf{ref}}(o\mymid q)\pi_{\mathsf{hp}}^{2\beta_{\mathsf{tts}}^{-1}}(o\mymid q)+\frac{\pi_{\mathsf{max}}^{2\beta_{\mathsf{tts}}^{-1}}\log^{2}N}{N^{2}}\right)\leq O\left(\frac{\pi_{\mathsf{max}}^{2\beta_{\mathsf{tts}}^{-1}}\log N}{N}\right)
\end{align}
with probability at least $1-O(N^{-10})$, 
where the last relation follows since $\sum_{o}\pi_{\mathsf{ref}}(o\mymid q)=1$. 
Now, define
\begin{align*}
\delta(q,o)
&\coloneqq \frac{1}{N}\pi_{\mathsf{hp}}^{\beta_{\mathsf{tts}}^{-1}}(o\mymid q) + \frac{1}{N}\sum_{i\ne 1} \pi_{\mathsf{hp}}^{\beta_{\mathsf{tts}}^{-1}}(o^{(i)}\mymid q) - \sum_{o} \pi_{\mathsf{ref}}(o\mymid q)\pi_{\mathsf{hp}}^{\beta_{\mathsf{tts}}^{-1}}(o\mymid q).
\end{align*}
The bound \eqref{eq:bernstein} tells us that, with probability at least $1-O(N^{-10})$, 
\begin{align}\label{eq:bound-delta}
|\delta(q,o)|\le \frac{1}{N}\pi_{\mathsf{max}}^{\beta_{\mathsf{tts}}^{-1}} + \frac{N-1}{N}O\left(\sqrt{\frac{\log N}{N}}\pi_{\mathsf{max}}^{\beta_{\mathsf{tts}}^{-1}}\right) + O\left( \frac{1}{N}\pi_{\mathsf{max}}^{\beta_{\mathsf{tts}}^{-1}} \right)
\leq O\left( 
\sqrt{\frac{\log N}{N}}\pi_{\mathsf{max}}^{\beta_{\mathsf{tts}}^{-1}} \right).
\end{align} 
%
%
 %
Substitution into \eqref{eq:testtimescaling} yields that: for any $N\gtrsim \exp(2r_{\mathsf{max}}\beta_{\mathsf{tts}}^{-1})\log N$, 
\begin{align*}
\mathbb{P}(o^{\mathsf{best}} = o\mymid q) 
&= \mathbb{E}\left[\frac{N\pi_{\mathsf{ref}}(o\mymid q)\pi_{\mathsf{hp}}^{\beta_{\mathsf{tts}}^{-1}}(o\mymid q)}{\pi_{\mathsf{hp}}^{\beta_{\mathsf{tts}}^{-1}}(o\mymid q) + \sum_{i\ne 1} \pi_{\mathsf{hp}}^{\beta_{\mathsf{tts}}^{-1}}(o^{(i)}\mymid q)}\right]\notag\\
&=\mathbb{E}\left[\frac{\sum_{o} \pi_{\mathsf{ref}}(o\mymid q)\pi_{\mathsf{hp}}^{\beta_{\mathsf{tts}}^{-1}}(o\mymid q)}{\sum_{o} \pi_{\mathsf{ref}}(o\mymid q)\pi_{\mathsf{hp}}^{\beta_{\mathsf{tts}}^{-1}}(o\mymid q)+\delta}\cdot \frac{\pi_{\mathsf{ref}}(o\mymid q)\pi_{\mathsf{hp}}^{\beta_{\mathsf{tts}}^{-1}}(o\mymid q)}{\sum_{o} \pi_{\mathsf{ref}}(o\mymid q)\pi_{\mathsf{hp}}^{\beta_{\mathsf{tts}}^{-1}}(o\mymid q)}\right]\notag\\
&=\mathbb{E}\left[\left(1-\frac{\delta}{\sum_{o} \pi_{\mathsf{ref}}(o\mymid q)\pi_{\mathsf{hp}}^{\beta_{\mathsf{tts}}^{-1}}(o\mymid q)+\delta}\right)\frac{\pi_{\mathsf{ref}}(o\mymid q)\pi_{\mathsf{hp}}^{\beta_{\mathsf{tts}}^{-1}}(o\mymid q)}{\sum_{o} \pi_{\mathsf{ref}}(o\mymid q)\pi_{\mathsf{hp}}^{\beta_{\mathsf{tts}}^{-1}}(o\mymid q)}\right]\notag\\
&\overset{\text{(i)}}{=} \left(1 + O\left(\sqrt{\frac{\log N}{N}}\frac{\pi_{\mathsf{max}}^{\beta_{\mathsf{tts}}^{-1}}}{\pi_{\mathsf{min}}^{\beta_{\mathsf{tts}}^{-1}}}\right) + O\left(\frac{1}{N^{10}}\frac{\pi_{\mathsf{max}}^{\beta_{\mathsf{tts}}^{-1}}}{\pi_{\mathsf{min}}^{\beta_{\mathsf{tts}}^{-1}}}\right)\right)\frac{\pi_{\mathsf{ref}}(o\mymid q)\pi_{\mathsf{hp}}^{\beta_{\mathsf{tts}}^{-1}}(o\mymid q)}{\sum_{o} \pi_{\mathsf{ref}}(o\mymid q)\pi_{\mathsf{hp}}^{\beta_{\mathsf{tts}}^{-1}}(o\mymid q)} \notag\\
&\overset{\text{(ii)}}{=}  \bigg(1 + O\Big(\frac{\exp(r_{\mathsf{max}}/\beta_{\mathsf{tts}})\sqrt{\log N}}{\sqrt{N}}\Big)\bigg)\frac{\pi_{\mathsf{ref}}(o\mymid q)\pi_{\mathsf{hp}}^{\beta_{\mathsf{tts}}^{-1}}(o\mymid q)}{\sum_{o} \pi_{\mathsf{ref}}(o\mymid q)\pi_{\mathsf{hp}}^{\beta_{\mathsf{tts}}^{-1}}(o\mymid q)},
\end{align*}
where (i) follows from the fact that \eqref{eq:bound-delta} holds with probability at least $1-O(N^{-10})$, as well as the basic inequality 
\begin{align*}
\left|\frac{\delta}{\sum_{o} \pi_{\mathsf{ref}}(o\mymid q)\pi_{\mathsf{hp}}^{\beta_{\mathsf{tts}}^{-1}}(o\mymid q)+\delta}\right| 
\le
\begin{cases}
\frac{N\pi_{\mathsf{max}}^{\beta_{\mathsf{tts}}^{-1}}}{\pi_{\mathsf{hp}}^{\beta_{\mathsf{tts}}^{-1}}(o\mymid q) + \sum_{i\ne 1} \pi_{\mathsf{hp}}^{\beta_{\mathsf{tts}}^{-1}}(o^{(i)}\mymid q)} \le \frac{\pi_{\mathsf{max}}^{\beta_{\mathsf{tts}}^{-1}}}{\pi_{\mathsf{min}}^{\beta_{\mathsf{tts}}^{-1}}}, & \text{if }\delta<0, \\
1, & \text{if }\delta\geq 0.
\end{cases}
\end{align*}
and (ii) uses $\pi_{\mathsf{max}}/\pi_{\mathsf{min}}\le \exp(r_{\mathsf{max}})$ as asserted in \eqref{eq:best-of-N-assumption-pi}.
As a consequence, when $N\to\infty$, the  distribution induced by this form of test time scaling converges to
\begin{align}
\mathbb{P}(o^{\mathsf{best}} = o\mymid q) \overset{N\rightarrow \infty}{\longrightarrow} 
\frac{\pi_{\mathsf{ref}}(o\mymid q)\pi_{\mathsf{hp}}^{\beta_{\mathsf{tts}}^{-1}}(o\mymid q)}{\sum_{o} \pi_{\mathsf{ref}}(o\mymid q)\pi_{\mathsf{hp}}^{\beta_{\mathsf{tts}}^{-1}}(o\mymid q)}.
\end{align}

\section{Diffusion models}

In this section, we shift our focus to to diffusion models, which represent a different data generation paradigm compared to the autoregressive model. Interestingly,  
a few approaches proposed in Section~\ref{sec:AR-model} for the AR model can be extended to accommodate diffusion models.


 In this section, we consider the task of conditional sampling: given condition $c$, one is asked to generate a novel data instance (e.g., an image, a video) whose distribution resembles a target data distribution $p_{\mathsf{data}}$.  We shall present our findings after introducing some preliminary background. 

\subsection{Preliminaries}

%

\paragraph{Score learning.} 
Conditional diffusion models rely on learning the conditional (Stein) score function from training data drawn from $p_{\mathsf{data}}$ (in conjunction with conditioning information). While we do not delve into details here (as these will not be used in our results), we note that score learning typically involves minimizing the following objective: 
\begin{align}\label{eq:score-training}
\mathop{\text{minimize}}\limits_{\theta}~~\mathop{\mathbb{E}}\limits_{x_0\sim p_{\mathsf{data}}, \varepsilon\sim\mathcal{N}(0,I), t}\big[\|\varepsilon - \varepsilon_{\theta}(x_t\mymid c, t)\|_2^2\big]. 
\end{align}
In this expression,  
$x_t = \sqrt{1-\sigma_t^2}x_0 + \sigma_t\varepsilon$, where $x_0\sim p_{\mathsf{data}}$ denotes a data instance sampled from the target distribution,  $\varepsilon\sim \mathcal{N}(0,I)$ is the additive Gaussian noise, and $\sigma_t>0$ is some prescribed coefficient dictating the standard deviation of the injected noise at time $t$. 
Intuitively, 
\eqref{eq:score-training} can be interpreted as predicting the additive noise using an ML model $\varepsilon_{\theta}$, given a noisy data sample at time $t$ and conditioned on $c$.  

\paragraph{Classifier-free diffusion guidance.} 

As a widely adopted technique for diffusion-based conditional sampling, classifier-free guidance attempts to generate samples from a distribution proportional to \citep{ho2021classifier,dhariwal2021diffusion}
\begin{align}
\label{eq:CFG-goal}
p_{\mathsf{data}}(c\mymid x)^{w_{\mathsf{cfg}}}p_{\mathsf{data}}(x\mymid c),
\end{align}
where $p_{\mathsf{data}}(c\mymid x)$ is raised to some prescribed power $w_{\mathsf{cfg}}>0$ to sharpen the classifier probability.  
Informally, in view of the Bayes rule  $p_{\mathsf{data}}(c\mymid x)\propto p_{\mathsf{data}}(x\mymid c)/p_{\mathsf{data}}(x)$, the ratio $p_{\mathsf{data}}(x\mymid c)/p_{\mathsf{data}}(x)$ is now sharpened to be proportional to 
$$
\frac{p_{\mathsf{data}}(c\mymid x)^{w_{\mathsf{cfg}}}p_{\mathsf{data}}(x\mymid c)}{p_{\mathsf{data}}(x)}~\propto~ \bigg(\frac{p_{\mathsf{data}}(x\mymid c)}{p_{\mathsf{data}}(x)}\bigg)^{1+w_{\mathsf{cfg}}}.$$
However, as noted in \citet{bradley2024classifier},  the standard classifier-free guidance approach falls short of producing samples from the target distribution $p_{\mathsf{data}}(c\mymid x)^{w_{\mathsf{cfg}}}p_{\mathsf{data}}(x\mymid c)$, since the effective score $(1+w_{\mathsf{cfg}})\varepsilon_\theta(x_t\mymid c,t)-\varepsilon_\theta(x_t\mymid t)$ adopted in classifier-free guidance does not necessarily correspond to a valid diffusion forward process. See \citet{bradley2024classifier} for more precise discussions.   


\paragraph{Reward-directed diffusion models.} This approach improves sample quality in diffusion models by leveraging a reward model,  which ideally captures human preferences about the generated samples and may be trained on a dataset of human evaluations  \citep{black2023training,gao2024reward,fan2023dpok}.  
Given some reward model $r(x, c)$, reward-directed diffusion models seek to generate samples that maximize the expected reward while,  in the meantime, remaining close to the data distribution $p_{\mathsf{data}}(\cdot\mymid c)$.  More precisely, it aims to solve the following KL-regularized problem: 
\begin{align}
\mathop{\text{maximize}}\limits_{\theta} ~~\mathop{\mathbb{E}}\limits_{x \sim p_{\theta}(\cdot\mymid c)}[r(x, c)] - \beta_{\mathsf{rd}}\mathsf{KL}\big(p_{\theta}(\cdot\mymid c) \parallel p_{\mathsf{data}}(\cdot\mymid c)\big),
\label{eq:reward-directed-diffusion}
\end{align}
where $\beta_{\mathsf{rd}}>0$ is some regularization parameter. 
Clearly, this is similar in spirit to the RLHF objective \eqref{eq:obj-RLHF-0}. 

Nevertheless, in stark contrast to the AR model setting, diffusion models do not learn to represent the distribution $p_{\theta}(\cdot\mymid c)$  explicitly, but rather only perform data generation with the aid of the socre function estimate $\varepsilon_\theta(x\mymid c,t)$.
Consequently, the KL divergence in \eqref{eq:reward-directed-diffusion} cannot be computed directly. 
To address this issue, one possible solution is to approximate the KL divergence of interest using the following training loss:
\begin{align}
\mathsf{KL}(p_{\theta}(\cdot\mymid c) \parallel p_{\mathsf{data}}(\cdot\mymid c)) \approx \mathbb{E}_{x_0, \varepsilon, t}\big[\|\varepsilon - \varepsilon_{\theta}(x_t\mymid c, t)\|_2^2 \big],
\label{eq:KL-approx-diffusion}
\end{align}
which converts the objective \eqref{eq:reward-directed-diffusion} into
\begin{align}
\mathop{\text{maximize}}\limits_{\theta} ~~\mathop{\mathbb{E}}\limits_{x \sim p_{\theta}(\cdot\mymid c)}[r(x, c)] - \beta_{\mathsf{rd}}\mathbb{E}_{x_0, \varepsilon, t}\big[\|\varepsilon - \varepsilon_{\theta}(x_t\mymid c, t)\|_2^2 \big],
\label{eq:reward-directed-diffusion-eps}
\end{align}
Note, however, that this approximation \eqref{eq:KL-approx-diffusion} is only guaranteed to be a valid lower bound \citep{li2024score}, which still does not preclude the possibility of being inaccurate.

\subsection{Connection between test-time scaling and diffusion guidance}

Considering the shared goal of the AR model and diffusion model, we  now extend the test-time scaling technique studied in Section~\ref{sec:AR-RL-tts} to the setting of diffusion model.

Assuming access to $p_\mathsf{data}(c\mymid x)$, 
one can improve sample quality via the following test-time scaling technique --- i.e., a soft variation of the best-of-$N$ method:
\begin{itemize}
\item[(i)] 
Generate $N$ independent candidate samples $x^{(1)},\ldots,x^{(N)}\sim p_{\mathsf{data}}(\cdot\mymid c)$;

\item[(ii)] Select $x^{\mathsf{best}}$ from these candidates using the stochastic strategy:
\begin{align*}
\mathbb{P}(x^{\mathsf{best}} = x^{(n)}\mymid c, x^{(1)}, x^{(2)}, \ldots, x^{(N)}) \propto p_{\mathsf{data}}^{w_{\mathsf{tts}}}(c\mymid x^{(n)})
\end{align*}
for some hyperparameter $w_{\mathsf{tts}}>0$. Clearly, when $w_{\mathsf{tts}}\rightarrow \infty$, this approach effectively selects the sample among the $N$ candidates that maximizes the classifier probability $p_{\mathsf{data}}(c\mymid \cdot)$. 
\end{itemize}

Repeating the same arguments as in Sections~\ref{sec:AR-RL-tts} and \ref{sec:proof-AR-model} (which we omit here for brevity), we can readily see that: 
as $N\to\infty$, the distribution of the selected sample $x^{\mathsf{best}}$ converges to
\begin{align}
p_{x^{\mathsf{best}}}(x\mymid c) 
\overset{N\rightarrow \infty}{\longrightarrow} \frac{p_{\mathsf{data}}^{w_{\mathsf{tts}}}(c\mymid x)p_{\mathsf{data}}(x\mymid c)}{\int p_{\mathsf{data}}^{w_{\mathsf{tts}}}(c\mymid x') p_{\mathsf{data}}(x'\mymid c)\mathrm{d} x'},
\qquad \text{for any }x, c.
\end{align}
Consequently, this test-time scaling approach achieves, asymptotically, the aim of classifier-free guidance in approaching the sharpened target distribution \eqref{eq:CFG-goal}, provided that $w_{\mathsf{cfg}}=w_{\mathsf{tts}}$.

\begin{remark}[Estimation of $p_{\mathsf{data}}(c\mymid x)$]
A key step of the above test-time scaling approach lies in estimating the classifier probability $p_\mathsf{data}(c\mymid x)$.
According to \citet[Eqn.~(4.6)]{li2024score}, it can be approximated by
\begin{align}
p_{\mathsf{data}}(c\mymid x) &\propto\frac{p_{\mathsf{data}}(x\mymid c)}{p_{\mathsf{data}}(x) }\approx \exp\big(-L(x;c)\big),
\end{align}
where
\begin{align}
L(x;c) \coloneqq \mathop{\mathbb{E}}\limits_{\varepsilon\sim\mathcal{N}(0,I),t\sim p_t}\Big[\left\|\varepsilon - \varepsilon_\theta(x_t\mymid c, t)\right\|_2^2-\left\|\varepsilon - \varepsilon_{\theta}(x_t\mymid t)\right\|_2^2 \mid x_t = \sqrt{1-\sigma_t^2}x + \sigma_t \varepsilon\Big],
\end{align}
and the distribution of $p_t$ depends on the variance of injected noise in diffusion models.
In particular, $p_t\propto \frac{\sigma_{t+1}^2-\sigma_{t}^2}{2\sigma_t^2(1-\sigma_t^2)}$ for $t=1,\ldots, T$ and $\sigma_1\le \ldots \sigma_T$.
\end{remark}

\subsection{An alternative resampling approach for reward-directed diffusion models}

As alluded to previously, one can sometimes improve the conditional sampling quality of diffusion models via reward-directed techniques (cf.~\eqref{eq:reward-directed-diffusion} and \eqref{eq:reward-directed-diffusion-eps}),  effectively  adjusting   the distribution $p_{\mathsf{data}}(x\mymid c)$ to $\exp(r(x,c)/\beta_{\mathsf{rd}})p_{\mathsf{data}}(x\mymid c)$ based on a given reward model $r$.


This insight inspires the development of an RL-free resampling approach to tackling the reward-directed objective \eqref{eq:reward-directed-diffusion-eps}, akin to the method proposed for the AR model in Section~\ref{sec:resampling-AR}. 
Specifically,  we aim to directly train a new score function estimate $\widetilde{\varepsilon}_\theta(\cdot\mymid c,t)$ to approximate the gradient of the log-density of $\widetilde{X}_0 + \sigma_t\varepsilon$, where $\widetilde{X}_0\sim \exp(r(\cdot, c)/\beta_{\mathsf{rd}}) p_{\mathsf{data}}(\cdot\mymid c)$ and $\varepsilon \sim \mathcal{N}(0,I)$.
Similar to \eqref{eq:fintunenoRL-AR}, we propose to accomplish this via the following training objective:
\begin{align}\label{eq:fintunenoRL-DF}
\mathop{\text{maximize}}\limits_{\theta} ~~Z_{\mathsf{rd}}^{-1}(c)\mathop{\mathbb{E}}\limits_{x \sim p_{\mathsf{data}}(\cdot\mymid c),\varepsilon,t}\left[\exp\big(r(x, c)/\beta_{\mathsf{rd}}\big)\left\|\varepsilon - \widetilde{\varepsilon}_{\theta}(x_t\mymid c, t)\right\|_2^2\right],
\end{align}
which is equivalent to solving
\begin{align}\label{eq:resampling-reward-directed}
\mathop{\text{maximize}}\limits_{\theta} \mathop{\mathbb{E}}\limits_{x \sim Z_{\mathsf{rd}}^{-1}(c)\exp(r(\cdot, c)/\beta_{\mathsf{rd}})p_{\mathsf{data}}(\cdot\mymid c),\varepsilon,t}\left[\big\|\varepsilon - \widetilde{\varepsilon}_{\theta}(x_t\mymid c, t)\big\|_2^2\right].
\end{align}
Here, the normalized constant is given by
\begin{align*}
Z_{\mathsf{rd}}(c) = \mathop{\mathbb{E}}\limits_{x \sim p_{\mathsf{data}}(\cdot\mymid c)}\big[\exp\big(r(x, c)/\beta_{\mathsf{rd}}\big)\big].
\end{align*}
As before, this empirical risk minimization objective \eqref{eq:resampling-reward-directed} can be implemented in practice by drawing data samples from $p_{\mathsf{data}}(\cdot\mymid c)$ and applying importance sampling, without resorting to any RL-based techniques.



\section{Conclusion}

In this note, we have explored several widely used post-training techniques, and have identified some intimate connections among them. In particular, we have drawn parallels between RLHF, RLIF, and test-time scaling, while also elucidating the intrinsic connections between diffusion guidance and test-time scaling. 
These insights have also inspired a resampling approach to alignment and reward-directed diffusion models without relying on explicit RL-based techniques. 




\section*{Acknowledgments}

G.~Li is supported in part by the Chinese University of Hong Kong Direct Grant for Research and the Hong Kong Research Grants Council ECS 2191363.
Y.~Chen is supported in part by the Alfred P.~Sloan Research Fellowship,  the ONR grants N00014-22-1-2354 and N00014-25-1-2344,  the NSF grants 2221009 and 2218773, 
the Wharton AI \& Analytics Initiative's AI Research Fund, 
and the Amazon Research Award.

\bibliographystyle{apalike}
\bibliography{bibfileDF}

\begin{thebibliography}{}

\bibitem[Beck, 2017]{beck2017first}
Beck, A. (2017).
\newblock {\em First-order methods in optimization}.
\newblock SIAM.

\bibitem[Black et~al., 2023]{black2023training}
Black, K., Janner, M., Du, Y., Kostrikov, I., and Levine, S. (2023).
\newblock Training diffusion models with reinforcement learning.
\newblock {\em arXiv preprint arXiv:2305.13301}.

\bibitem[Bradley and Nakkiran, 2024]{bradley2024classifier}
Bradley, A. and Nakkiran, P. (2024).
\newblock Classifier-free guidance is a predictor-corrector.
\newblock {\em arXiv preprint arXiv:2408.09000}.

\bibitem[Bradley and Terry, 1952]{bradley1952rank}
Bradley, R.~A. and Terry, M.~E. (1952).
\newblock Rank analysis of incomplete block designs: I. the method of paired comparisons.
\newblock {\em Biometrika}, 39(3/4):324--345.

\bibitem[Brown et~al., 2024]{brown2024large}
Brown, B., Juravsky, J., Ehrlich, R., Clark, R., Le, Q.~V., R{\'e}, C., and Mirhoseini, A. (2024).
\newblock Large language monkeys: Scaling inference compute with repeated sampling.
\newblock {\em arXiv preprint arXiv:2407.21787}.

\bibitem[Chen et~al., 2021]{chen2021spectral}
Chen, Y., Chi, Y., Fan, J., Ma, C., et~al. (2021).
\newblock Spectral methods for data science: A statistical perspective.
\newblock {\em Foundations and Trends{\textregistered} in Machine Learning}, 14(5):566--806.

\bibitem[Dhariwal and Nichol, 2021]{dhariwal2021diffusion}
Dhariwal, P. and Nichol, A. (2021).
\newblock Diffusion models beat {GAN}s on image synthesis.
\newblock {\em Advances in neural information processing systems}, 34:8780--8794.

\bibitem[Fan et~al., 2023]{fan2023dpok}
Fan, Y., Watkins, O., Du, Y., Liu, H., Ryu, M., Boutilier, C., Abbeel, P., Ghavamzadeh, M., Lee, K., and Lee, K. (2023).
\newblock {DPOK}: Reinforcement learning for fine-tuning text-to-image diffusion models.
\newblock {\em Advances in Neural Information Processing Systems}, 36:79858--79885.

\bibitem[Gao et~al., 2024]{gao2024reward}
Gao, X., Zha, J., and Zhou, X.~Y. (2024).
\newblock Reward-directed score-based diffusion models via {Q}-learning.
\newblock {\em arXiv preprint arXiv:2409.04832}.

\bibitem[Ho and Salimans, 2021]{ho2021classifier}
Ho, J. and Salimans, T. (2021).
\newblock Classifier-free diffusion guidance.
\newblock In {\em NeurIPS 2021 Workshop on Deep Generative Models and Downstream Applications}.

\bibitem[Jurafsky and Martin, 2025]{jm3}
Jurafsky, D. and Martin, J.~H. (2025).
\newblock {\em Speech and Language Processing: An Introduction to Natural Language Processing, Computational Linguistics, and Speech Recognition with Language Models}.
\newblock 3rd edition.

\bibitem[Kang et~al., 2024]{kang2024unfamiliar}
Kang, K., Wallace, E., Tomlin, C., Kumar, A., and Levine, S. (2024).
\newblock Unfamiliar finetuning examples control how language models hallucinate.
\newblock {\em arXiv preprint arXiv:2403.05612}.

\bibitem[Kang et~al., 2025]{kang2025scalable}
Kang, Z., Zhao, X., and Song, D. (2025).
\newblock Scalable {best-of-$N$} selection for large language models via self-certainty.
\newblock {\em arXiv preprint arXiv:2502.18581}.

\bibitem[Kaufmann et~al., 2023]{kaufmann2023survey}
Kaufmann, T., Weng, P., Bengs, V., and H{\"u}llermeier, E. (2023).
\newblock A survey of reinforcement learning from human feedback.
\newblock {\em arXiv preprint arXiv:2312.14925}.

\bibitem[Li and Yan, 2024]{li2024score}
Li, G. and Yan, Y. (2024).
\newblock A score-based density formula, with applications in diffusion generative models.
\newblock {\em arXiv preprint arXiv:2408.16765}.

\bibitem[Muennighoff et~al., 2025]{muennighoff2025s1}
Muennighoff, N., Yang, Z., Shi, W., Li, X.~L., Fei-Fei, L., Hajishirzi, H., Zettlemoyer, L., Liang, P., Cand{\`e}s, E., and Hashimoto, T. (2025).
\newblock s1: Simple test-time scaling.
\newblock {\em arXiv preprint arXiv:2501.19393}.

\bibitem[Ouyang et~al., 2022]{ouyang2022training}
Ouyang, L., Wu, J., Jiang, X., Almeida, D., Wainwright, C., Mishkin, P., Zhang, C., Agarwal, S., Slama, K., Ray, A., et~al. (2022).
\newblock Training language models to follow instructions with human feedback.
\newblock {\em Advances in neural information processing systems}, 35:27730--27744.

\bibitem[Snell et~al., 2024]{snell2024scaling}
Snell, C., Lee, J., Xu, K., and Kumar, A. (2024).
\newblock Scaling {LLM} test-time compute optimally can be more effective than scaling model parameters.
\newblock {\em arXiv preprint arXiv:2408.03314}.

\bibitem[Verdun et~al., 2025]{verdun2025soft}
Verdun, C.~M., Oesterling, A., Lakkaraju, H., and Calmon, F.~P. (2025).
\newblock Soft best-of-n sampling for model alignment.
\newblock {\em arXiv preprint arXiv:2505.03156}.

\bibitem[Zhao et~al., 2025]{zhao2025learning}
Zhao, X., Kang, Z., Feng, A., Levine, S., and Song, D. (2025).
\newblock Learning to reason without external rewards.
\newblock {\em arXiv preprint arXiv:2505.19590}.

\bibitem[Ziegler et~al., 2019]{ziegler2019fine}
Ziegler, D.~M., Stiennon, N., Wu, J., Brown, T.~B., Radford, A., Amodei, D., Christiano, P., and Irving, G. (2019).
\newblock Fine-tuning language models from human preferences.
\newblock {\em arXiv preprint arXiv:1909.08593}.

\end{thebibliography}

\end{document}